\documentclass[runningheads]{llncs}
\usepackage{graphicx}
\usepackage{amsmath}
\usepackage{tabularx}
\begin{document}
\title{A Span Extraction Approach for Information Extraction on Visually-Rich Documents}
\titlerunning{A Span Extraction Approach for Information Extraction on VRDs}
%
\author{
    Tuan-Anh D. Nguyen\inst{1} \and
    Hieu M. Vu\inst{1} \and
    Nguyen Hong Son\inst{1} \and \\
    Minh-Tien Nguyen\inst{1,2}
}
\authorrunning{Tuan-Anh et. al.}
%
%
\institute{
    Cinnamon AI, 10th floor, Geleximco building, 36 Hoang Cau, Dong Da, \\Hanoi, Vietnam.\\
    \email{\{tadashi,ian,levi,ryan.nguyen\}@cinnamon.is}
\and
    Hung Yen University of Technology and Education, Hung Yen, Vietnam.\\
    \email{tiennm@utehy.edu.vn}
}

\maketitle              
\begin{abstract}
Information extraction (IE) for visually-rich documents (VRDs) has achieved SOTA performance recently thanks to the adaptation of Transformer-based language models, which shows the great potential of pre-training methods. In this paper, we present a new approach to improve the capability of language model pre-training on VRDs. Firstly, we introduce a new query-based IE model that employs span extraction instead of using the common sequence labeling approach. Secondly, to extend the span extraction formulation, we propose a new training task focusing on modelling the relationships among semantic entities within a document. This task enables target spans to be extracted recursively and can be used to pre-train the model or as an IE downstream task. Evaluation on three datasets of popular business documents (\textit{invoices, receipts}) shows that our proposed method achieves significant improvements compared to existing models. The method also provides a mechanism for knowledge accumulation from multiple downstream IE tasks.
\keywords{Information Extraction \and Span Extraction \and Pre-trained model \and Visually-rich document.}
\end{abstract}
\section{Introduction \label{intro}}
Information Extraction (IE) for visually-rich documents (VRDs) is different from plain text documents in many aspects. VRDs contain sparser textual content, have strictly defined visual structures, and complex layouts which do not present in plain text documents. Due to these characteristics, existing works on IE for VRDs mostly employ Computer Vision \cite{dang2019end} and/or graph-based \cite{Qian2018,Liu2019,Vedova2019} techniques. Recently, BERT provides more perspectives on VRDs from the Natural Language Processing point of view \cite{xu2020layoutlm,Xu2020LayoutLMv2MP,xu2021layoutxlm,hwang2020spatial,hong2021bros,Yu2020PICKPK}.

LayoutLM \cite{xu2020layoutlm} emerges as a simple yet potential approach, where the BERT architecture was extended by the addition of 2D positional embedding and visual embedding to its input embedding \cite{devlin2018bert}. The model was trained by the Masked Visual Language Modelling (MVLM) objective \cite{xu2020layoutlm}, which is also an adaptation of positional embedding and visual embedding to the Masked Language Modelling objective \cite{devlin2018bert} of BERT. More recently, LayoutLMv2 \cite{Xu2020LayoutLMv2MP} was introduced with more focus on better leveraging 2D positional and visual information.
There is also the addition of two new pre-training objectives that were designed for modeling image-text correlation.

While LayoutLM is very efficient to model multi-modal data like visually-rich documents, it bears a few drawbacks regarding practicality. First, pre-training a language model requires a large amount of data, and unlike plain text documents, visually-rich document data is not readily available in such large quantities, especially for low-resource languages (e.g. Japanese). Second, previous methods mostly utilize sequence labeling for IE, but we argue that this method might not work well for short entities due to the imbalance of classes and the training loss function \cite{10.1145/1597735.1597754,li2019dice}. Third, sequence labeling requires a fixed and predefined set of entity tags/classes for each dataset, which hinders the application of applying the same IE model to multiple datasets.

To address the aforementioned problems, we propose a new method employing span extraction instead of using sequence labeling. The method combines QA-based span extraction and multi-value extraction using a novel recursive relation predicting scheme. We also introduce a pre-training procedure that extends the LayoutLM to both English and non-English datasets by using semantic entities. Since the query is embedded independently with the context, this new method can be applied across multiple datasets, thus enable the seamless accumulation of knowledge through one common formulation.

\section{Information Extraction as Span Extraction \label{span_ie}}
The most common approach for extracting information from visually-rich documents is sequence labeling which assigns a label for each token of the document \cite{xu2020layoutlm,Xu2020LayoutLMv2MP,xu2021layoutxlm,hwang2020spatial,hong2021bros,Yu2020PICKPK}. Even though this solution showed promising results, we argue that the sequence labeling might not work well when some entity types have only a small amount of samples or when the target entities are short spans of text in dense documents. The possible reason is the imbalance of labels when training IE models. Moreover, sequence labeling requires an explicit definition of token classes, which diminishes the reusability of the model on other tasks.

To overcome this challenge, we follow the span extraction approach, also known as Extractive Question Answering (QA) in NLP \cite{devlin2018bert}. Different from sequence labeling which assigns a label for every token based on their embedding, QA models predict the start and the end positions of the corresponding answer for each query. For the IE problem, a query represents a required field/tag that we want to extract from the original document context. Each query is represented as a learnable embedding vector and kept separately from the main language model encoder.
\begin{figure*}[!h]
	\centering
	\includegraphics[width=.7\textwidth]{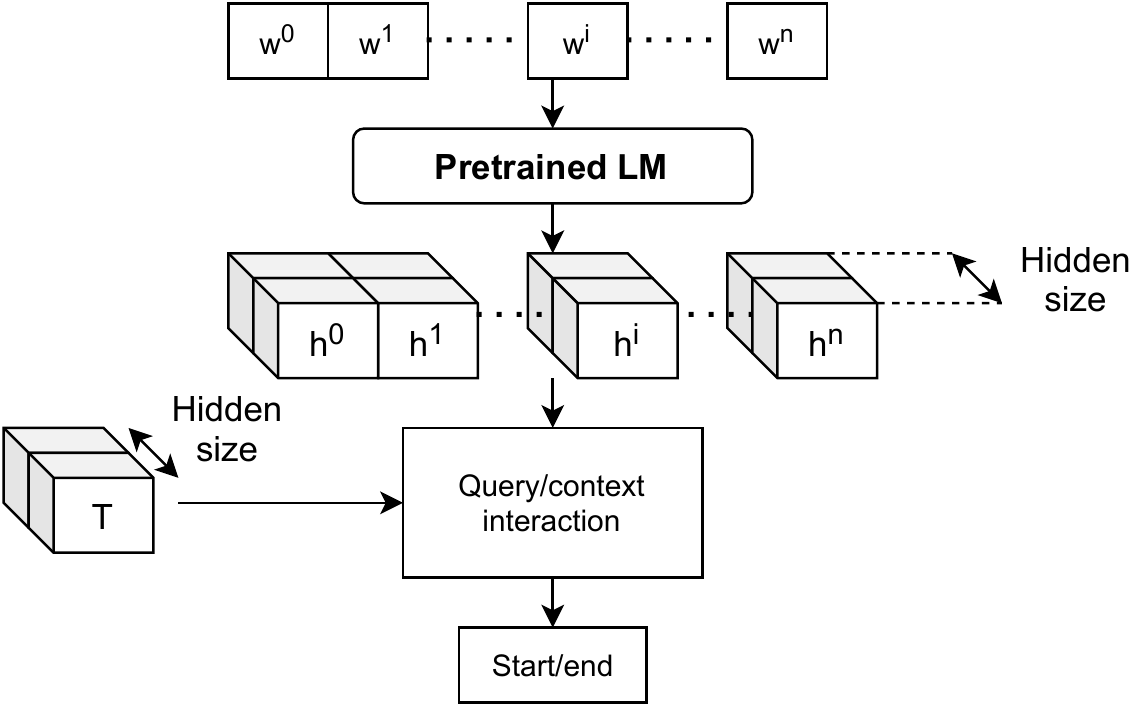}\vspace{-0.1cm}
	\caption{The span extraction model.}\label{fig:model}\vspace{-0.4cm}
\end{figure*}

Figure \ref{fig:model} describes our span extraction model. Let $D =\{w_{1}, w_{2},...,w_{n}\}$ is the input sequence and $H = [h^1, ..., h^j, ..., h^n] = f( D|\Theta)$ is the hidden representation of the input sequence produced by the pretrained encoder $f()$ with $h^{i} \in \mathbf{R}^c$ is the representation vector of the token $i^{th}$, $T \in \mathbf{R}^c$ is the vector representation of the query (field embedding) and $c$ is the hidden size of the encoder, the start and end position of the corresponding answer is calculated as:
\begin{align}
    \label{eq:span_ie}
    start, end = g(T, H)
\end{align}
with $g()$ is represented by an attention layer \cite{Luong2015EffectiveAT} that captures the interaction between the query $T$ and the context $H$.

To calculate the span extraction loss, the $softmax$ operation is applied over all tokens in the context instead of over the token classes for each token as in sequence tagging. The softmax loss on all sequence tokens eliminates the class imbalance problem for shorter answer entities.
The separation of question (query) embedding and context embedding also opens the potential of accumulating and reusing information on different datasets.

\section{Pre-training Objectives with Recursive Span Extraction \label{span_pretrain}}
\subsection{LayoutLM for Low-resource Languages}
\label{transfer_pretrain}

This section describes some effective methods for transferring the LayoutLM to low-resource languages, e.g. Japanese. Pre-training a language model from scratch with the MLM objective normally requires millions of data and can take a long time for training. While such an amount of data can be acquired in plain text, visually rich data does not naturally exist in large quantities due to the lack of word-level OCR annotations. Since pre-trained weights are available for English only, applying LayoutLM to other languages is still an open question. To overcome this, we propose a simple transferred pre-training procedure for LayoutLM that first takes advantage of the available pre-trained weights of BERT and then transfers its contextual representation to LayoutLM.

\subsubsection{Overcoming data shortage.} To transfer the LayoutLM model from English (the source language) to another language (the target language), we follow the following procedure. First, the model's word embedding layer was initialized from the pre-trained BERT of the target language and the rest layers (including positional embedding layers, encoder layers and the MVLM head) were initialized from the publicly available LayoutLM for English. After that, the model was trained with the MVLM objective using a much smaller dataset (about 17,000 samples) with 100 epochs. The final model can be fine-tuned to downstream tasks such as sequence labeling, document classification or question answering.

The word embedding layer maps plain text tokens into a semantically meaningful latent space, while the latent space can be similar, the mapping process itself is strictly distinct from language to language. We argue that spatial information is language-independent and can be useful among languages with alike reading orders. Similarly, the encoder layers are taken from the LayoutLM weights to leverage its ability to capture attention from both semantic and positional inputs. Our experiments on internal datasets show that this transferred pre-training setting yields significantly better results than without using the word embedding weights from BERT. Additionally, our dataset with 17,000 samples is not sufficient to pre-train the LayoutLM from scratch, which would give an unfair comparison between using our proposed transferred pre-training procedure and training from scratch. Thus, we plan to verify our approach on the recently introduced XFUN dataset \cite{xu2021layoutxlm}.

\subsubsection{Overcoming inadequate annotations.} VRDs usually exist in image formats without any OCR annotations. Thus, to address the lack of word-level OCR annotations, we used an in-house OCR reader to extract line-level annotations. Then, word-level annotations were approximated by dividing the bounding box of each line proportionally to the length of the words in that line.

\subsection{Span Extraction Pre-training} \label{span_pretrain}

To address the limitation of training data requirement when pre-training the LayoutLM model, we present a new pre-training objective that re-uses existing entities' annotations from downstream IE tasks. Our pre-train task bases on a recursive span extraction formulation that extends the traditional QA method to extract multiple answer spans from a single query. 

\subsubsection{\textbf{Recursive span extraction.}} 
\begin{figure*}[!h]
	\centering
	\includegraphics[width=.8\textwidth]{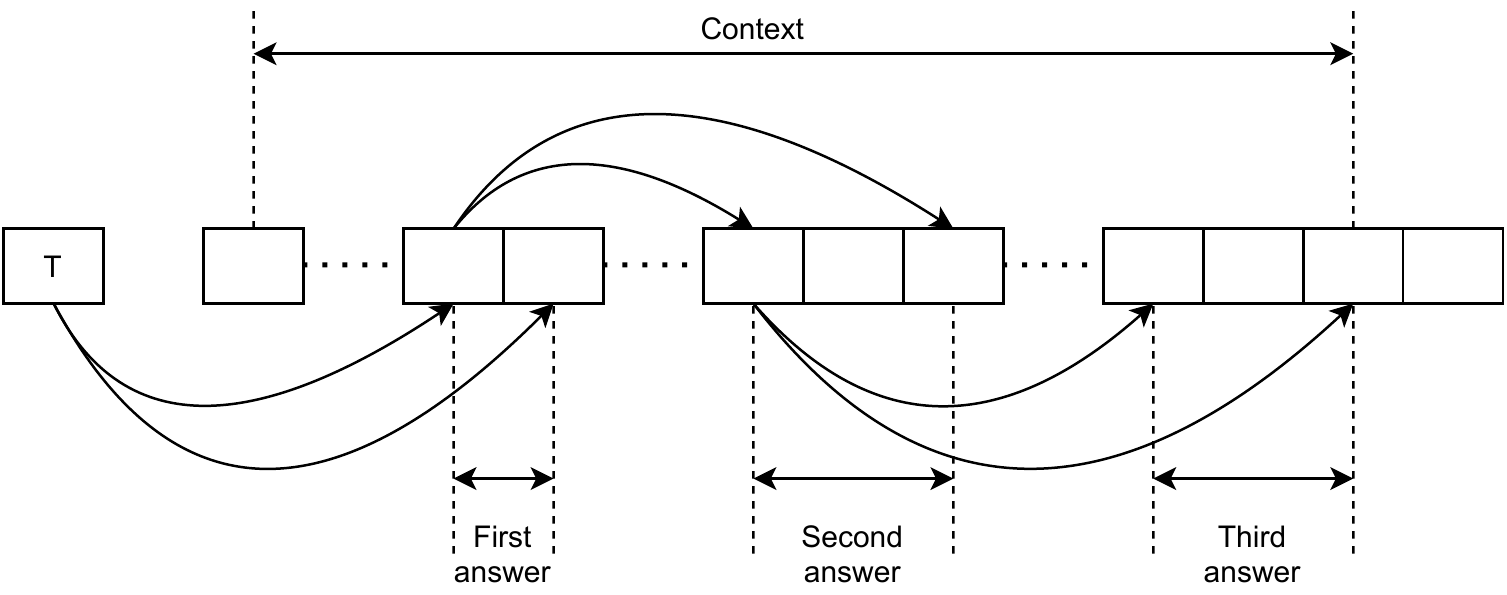}\vspace{-0.1cm}
	\caption{The recursive span extraction mechanism.}\label{fig:recusive_span}\vspace{-0.2cm}
\end{figure*}

As opposed to sequence labeling, our span extraction formulation allows a flexible independent query (field) with the context, which enables knowledge accumulation from multiple datasets by keeping updating a collection of question embeddings. However, one limitation of our span extraction model described in Fig. \ref{fig:model} is that the model can not extract more than one answer (entity) for each query, which is arguably crucial in many IE use cases. We, therefore, propose a strategy to extend the capability of the model by using a recursive link decoder mechanism which is described in Fig. \ref{fig:recusive_span}.

\begin{figure*}[!h]
	\centering
	\includegraphics[width=0.6\textwidth]{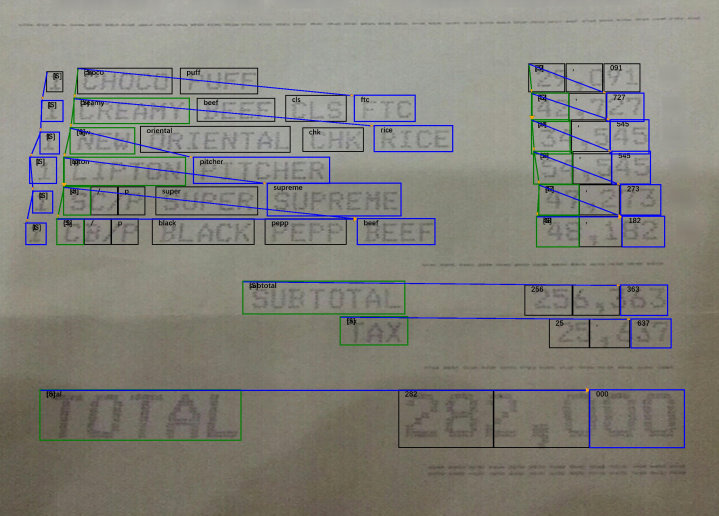}\vspace{-0.0cm}
	\caption{Recursive link prediction demonstrated on a receipt example \cite{park2019cord}. (\textit{green} and \textit{blue} arrows denote the target start and end positions of answer span corresponding to the query token (green bounding-boxes). The query tokens are the first token that starts each text segments. }\label{fig:link_prediction}\vspace{-0.4cm}
\end{figure*}

Using the same notation in Section \ref{span_ie}, our recursive mechanism is described as follows:

\begin{align}
    \label{eq:span_link_base}
    s_{1}, e_{1} &= g(T, H) \\
    \label{eq:span_link_recur}
    s_{i+1}, e_{i+1} &= g(h^{s_{i}}, H)
\end{align}

\noindent The first answer span is extracted by the vector representation $T$ of the query and the hidden representation $H$ of the context (Eq.~\ref{eq:span_link_base}). After that, the hidden representation corresponding to the start token of the $i$-th answer is used as the query vector to extract the $(i+1)$-th answer recursively (Eq.~\ref{eq:span_link_recur}). This recursive decoding procedure stops if $s_{i}=e_{i}=0$ or when $s_{i}= s_{j}, e_{i}=e_{j}$ with $j < i$. By using this method, we can decode all answer spans that belong to the same query/entity tag in a document context.

\subsubsection{\textbf{Pre-training with semantic entities.}} The recursive span extraction task forms a generic relation structure among entities, which reinforce the language model capability to represent context and spatial information in visually-rich documents. 
The model has to infer the continuous links between same-field items that often represented by vertical or horizontal links following the table structure.
A sample visualization of a document and the corresponding continuous linking formulation can be viewed in Fig. \ref{fig:link_prediction}. 

By forcing the model to reconstruct the relation between continuous entities which belong to the same query, we hypothesize that the model can produce better performance when being fine-tuned on downstream IE tasks. Experiment results in Section \ref{experiments} validate our hypothesis.

The proposed query-based architecture and recursive span decoding enable a generic IE formulation that is independent from the specific query tags of each document type. 
Recall that the query embedding was built separately from the main pre-trained model, the LayoutLM encoder will focus on modelling document context and interactions between its entities. 
We can pre-train the LayoutLM model seamlessly on multiple IE datasets with different key-value fields. 
To this end, we named this pre-training objective as \textit{Span extraction semantic pre-training}, which allows us to pretrain LayoutLM with several small-sized IE-annotated datasets in low-resource languages, instead of using large scale document data such as \cite{Xu2020LayoutLMv2MP}.

\section{Experiments  \label{experiments}}
\subsection{Experiment Setup}
\subsubsection{Dataset.}  
We use two internal self-collected datasets: one for pre-training the language model and one for evaluating IE performance as a downstream task. We also evaluate the performance of our proposed method on a public dataset consists of shop receipts \cite{park2019cord}, which represent a typical example of the document understanding task in commercial systems.
 Table \ref{tab:data} shows the statistics of datasets used in our experiments.
 
 \begin{table}[!h]
\centering
\setlength{\tabcolsep}{3pt}
\caption{\label{tab:data} 3 VRD document datasets. \textit{Italic} is internal.}\vspace{-0.1cm}
\begin{tabular}{lcccc}
\hline
\textbf{Data}  & \textbf{\#train} & \textbf{\#test} & \textbf{\#words/doc} & \textbf{\#fields}\\ \hline
\textit{Japanese Pre-training BizDocs}  & 5170 & - & 503 & 123 \\
\textit{Japanese Internal Invoice} & 700 & 300 & 421  & 25 \\
CORD  & 800 & 100 & 42 &  30  \\
\hline
\end{tabular} \vspace{-0.4cm}
\end{table}
 
 \paragraph{\textbf{Japanese pre-training BizDocs}} is a self-collected dataset. It consists of 5,170 document images and their OCR transcriptions. The dataset contains various types of business documents such as: financial report, invoice, medical receipt, insurance contract, \dots Each document type was annotated with an unique set of key-value fields which represent the important information that end-users want to extract. There are over 100 distinct fields and around 80,000 labeled entities. We use this dataset to pre-train our LayoutLM model with the proposed span-based objectives (Section \ref{span_pretrain}).
 
 \paragraph{\textbf{Japanese invoice}} is another Invoice documents dataset that disjoints with the aforementioned pre-training data. The data consists of 1,200 document images of shipping invoice and is annotated with 25 type of key-value fields. We divide the data into 700/200/300-sample sets that correspond the training/validation/testing data respectively.
 
 \paragraph{\textbf{CORD}} \cite{park2019cord} consists of Indonesian receipts collected from shops and restaurants. The dataset includes 800 receipts for training, 100 for validation, and 100 for testing. An image and a list of OCR annotations are provided alongside with each receipt. There are 30 fields which define with most important classes such as: store information, payment information, menu, total. The objective of the model is to label each semantic entity in the document and extract their values (Information Extraction) from input OCR annotations. The dataset is available publicly at \textit{\url{https://github.com/clovaai/cord}}.
 
\subsubsection{Training and inference.}
The English-based LayoutLM model starts from the pre-trained weight of \cite{xu2020layoutlm} which was provided with the paper, dubbed as \textit{EN-LayoutLM-base}. We used the \textit{LayoutLM-base} version from the original paper (113M parameters) due to hardware limitations. 
Another Japanese-based model (\textit{JP-LayoutLM-base}) was pre-trained by using the procedure described in Section \ref{transfer_pretrain} with the same configuration as \textit{LayoutLM-base}, which uses the MVLM task on our internal data. This model is intended to be used with Japanese datasets.

To conduct our experiments, we firstly pre-train the \textit{JP-LayoutLM-base} model on the \textit{Japanese Pre-training BizDocs} dataset with the proposed span-extraction objectives. 
Then, we perform fine-tuning on two IE datasets: \textit{Japanese invoice} and \textit{CORD} with the pre-trained weight \textit{JP-LayoutLM-base} and \textit{EN-LayoutLM-base} respectively. We compare the results of span-extraction with sequence labeling on these downstream IE tasks. 
Also, we measure the fine-tuning result with LayoutLM-base model without span-based pre-training to demonstrate the effect of the proposed method.
 
As for hyperparameters detail, the two models \textit{JP-LayoutLM-base} and \textit{EN-LayoutLM-base} share the same backbone with the hidden size of 768. In the Japanese model, we adopt the embedding layer and the vocabulary from the Japanese pre-trained BERT \texttt{cl-tohoku} \footnote{https://github.com/cl-tohoku/bert-japanese} which contains 32,000 sub-words. We also set the input sequence length to 512 in all tested models. The learning rate and the optimization method were kept $5e-5$ and Adam optimizer respectively \cite{xu2020layoutlm}. In term of loss function, with QA format, the standard cross entropy (CE) loss is used for start and end indices from all document tokens. With sequence tagging model, CE loss was used for the whole sequence classification output of the document.
 
 \subsubsection{Evaluation metrics.} We adopt the popular \textit{entity-level F1-score (micro)} for evaluation, which is commonly used in previous studies \cite{xu2020layoutlm,Xu2020LayoutLMv2MP,hwang2020spatial,hong2021bros}. The \textit{Entity F1-score (macro)} is used as the second metric to represent the overall result on all key-value fields, since it emphasizes model performance on some rare fields/tags. This situation is commonly appeared in industrial settings where data collection is costly for some infrequent fields.

\subsection{Results and Discussion}
\subsubsection{CORD.}
We report the results on the \textit{test} and \textit{dev} sub-sets of CORD in Table \ref{tab:cord_dev} and Table \ref{tab:cord_test} respectively. The span extraction method consistently improves the performance of IE in compare to existing sequence labeling method. We can observe $1\%$ - $3\%$ improvement in both metrics which is significant given competitive base result of default LayoutLM model. Interestingly, compared to the result from LayoutLMv2 \cite{Xu2020LayoutLMv2MP}, our model performs better than both \textit{LayoutLM-large} and \textit{LayoutLM-base-v2}, even though we only use a much lighter version of the model. This demonstrates the effectiveness of our span extraction formulation for downstream IE tasks. Note that the span extraction formulation is independent from the model architecture, thus we can expect the same improvement when starting from more complex pre-trained models. We leave this experiment as our future work.

Another finding from the results is that \textit{F1-score macro} is increased by an significant amount alongside with entity-level score (\textit{micro}) in both subsets. This illustrates that our method span-extraction helps to improve performance on all fields/tags, not only common tags that have more entities' annotation. 

\begin{table}[]
\centering
\caption{Results on the \textit{CORD (test) dataset} \\ \small{\textit{(*: result from original paper \cite{Xu2020LayoutLMv2MP})}}.}
\label{tab:cord_test}
\begin{tabular}{|l|c|c|c|}
\hline
\multicolumn{1}{|c|}{\textbf{Methods}} & \multicolumn{1}{c|}{\textbf{F1 (macro)}} & \multicolumn{1}{c|}{\textbf{F1 (micro)}} &
\multicolumn{1}{c|}{\textbf{\#Params}} \\ 
\hline \hline
EN-LayoutLM-base \textit{(seq labeling)}            & 80.08                   & 94.86         & 113M         \\ \hline
EN-LayoutLM-base \textit{(span extraction)}                   & \textbf{83.46}                   & \textbf{95.71} & 113M                 \\ \hline
EN-LayoutLM-large \textit{(seq labeling)} *                   & -                  & 94.93     & 343M              \\ \hline
EN-LayoutLM-base-v2 \textit{(seq labeling)} *                   & -             & 94.95   &  200M                  \\ \hline
\end{tabular}
\vspace{-0pt}
\end{table}


\begin{table}[]
\centering
\caption{Results on the \textit{CORD (dev) dataset}.}
\label{tab:cord_dev}
\begin{tabular}{|l|c|c|}
\hline
\multicolumn{1}{|c|}{\textbf{Methods}} & \multicolumn{1}{l|}{\textbf{F1-score (macro)}} & \multicolumn{1}{l|}{\textbf{F1-score (micro)}} \\ \hline \hline
EN-LayoutLM-base \textit{(seq labeling)}            & 80.74                   & 96.16                  \\ \hline
EN-LayoutLM-base \textit{(span extraction)}                   & \textbf{82.13}                   & \textbf{97.35}                  \\ \hline
\end{tabular}
\vspace{-0pt}
\end{table}

\subsubsection{Japanese invoice.}
Table \ref{tab:invoice} presents the result on \textit{Japanese invoice} data. We compare our method to sequence labeling, which shows similar improvements in term of both \textit{F1-score} metrics. Additionally, when adding the span-based pre-training task to the model, the performance of the downstream IE task increases by a substantial amount (+1.2\% in \textit{F1-score (micro)}). The improvement illustrates that our pre-training strategy can effectively improve the IE performance, given only semantic-annotated entities from other document types. 
This means that we can accumulate future IE datasets when fine-tuning as a continual learning scheme to further improve the base pre-train LayoutLM model. 

The \textit{F1-score (macro)} metric is also significantly improved when using span pre-training objectives compared to starting from the raw LayoutLM model. The observed phenomenon can be explained as the knowledge between different fields is effectively exploited to improve performance of less frequent tags, which often cause the \textit{macro} F1 metric to be lower than \textit{micro} in both datasets. The result aligns with our initial hypothesis in section \ref{transfer_pretrain}.

\begin{table}[]
\caption{Ablation study on the \textit{Japanese invoice dataset}}
\centering
\begin{tabular}{|l|c|c|}
\hline
\multicolumn{1}{|c|}{\textbf{Methods}} & \multicolumn{1}{l|}{\textbf{F1-score (macro)}} & \multicolumn{1}{l|}{\textbf{F1-score (micro)}} \\ \hline \hline
JP-LayoutLM-base \textit{(seq labeling)}             & 85.67                   & 90.15                  \\ \hline
JP-LayoutLM-base \textit{(span extraction)}             & 87.55                   & 91.34                  \\ \hline
\vtop{\hbox{\strut{JP-LayoutLM-base}}\hbox{\strut{ + \textbf{span pre-training} (\textit{seq labeling})}}}      & 88.02                   & 91.84                  \\ \hline
\strut  \vtop{\hbox{\strut{JP-LayoutLM-base}}\hbox{\strut{ + \textbf{span pre-training} (\textit{span extraction})}}}   & \textbf{89.76}                   & \textbf{92.55}                  \\ \hline
\end{tabular}
\label{tab:invoice}
\vspace{-17pt}
\end{table}

\subsubsection{Visualization.}
\begin{figure*}[!h]
	\centering
	\includegraphics[width=1.0\textwidth]{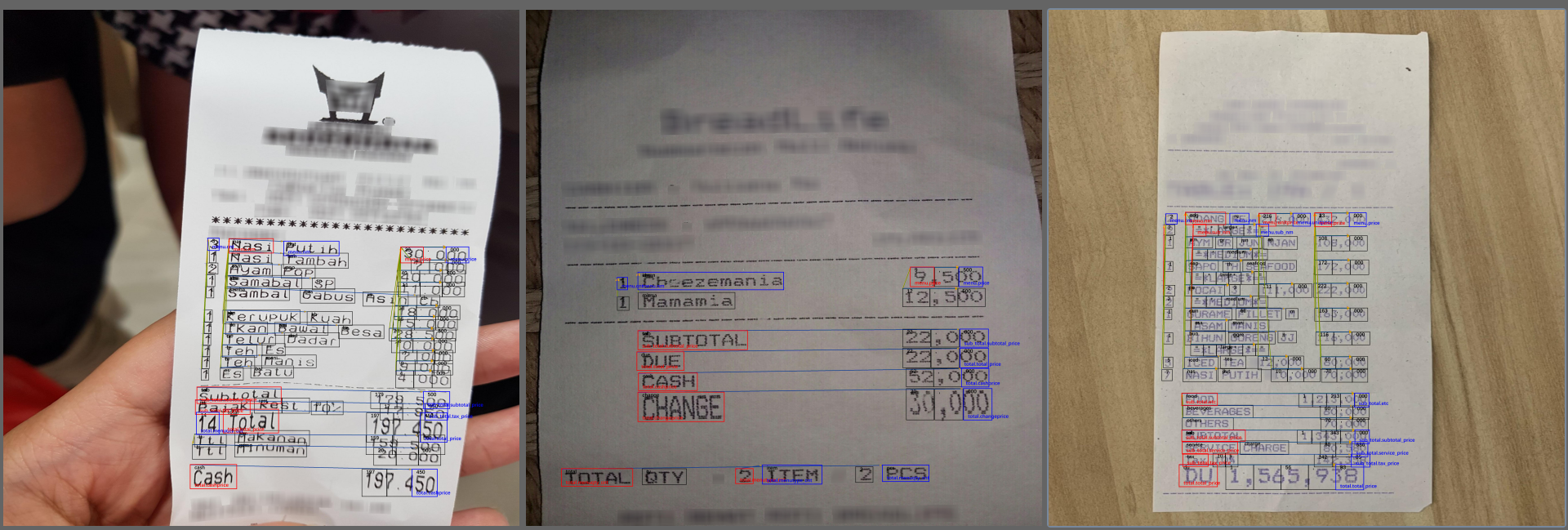}\vspace{-0.0cm}
	\caption{Visualization of model output on the \textit{CORD (test)} dataset \cite{park2019cord}. (\textit{red} and \textit{blue} boxes denote the output start and end positions of answer spans corresponding to each target field. Arrows represent the links between tokens.}\label{fig:cord_output}\vspace{-0.4cm}
\end{figure*}

The output of the model on the CORD (test) dataset can be visualized at Fig \ref{fig:cord_output}. The model can successfully infers spatial relations between various document elements to form the final IE output. We can observe some irregular structures of the table (due to camera-capture condition) which can be processed correctly. It demonstrates model capability to understand general layout structure of input documents.

\section{Conclusion} \label{conclusion}
This paper introduces a new span extraction formulation for the information extraction task on visually-rich documents. A pre-training scheme based on semantic entities annotation from IE data is also presented. The proposed method shows promising results on two IE datasets of business documents. Our method consistently achieves 1\%-3\% improvement compared to the sequence labeling approach, even with a smaller LayoutLM backbone. Furthermore, it can be applied to most existing pre-trained models for IE and opens a new direction to continuously improve the performance of pre-trained models through downstream tasks fine-tuning. For future work, we would like to consider the integration of our span extraction mechanism with the latest LayoutLMv2 model. We also plan to extend the model pre-training task with more diverse document relations such as: table structures, header-paragraph relations to further push the performance of pre-trained LM on VRDs. 
%
%
%
%
\bibliographystyle{splncs04}
\bibliography{ref}

\begin{thebibliography}{10}
\providecommand{\url}[1]{\texttt{#1}}
\providecommand{\urlprefix}{URL }
\providecommand{\doi}[1]{https://doi.org/#1}

\bibitem{dang2019end}
Dang, T.A.N., Thanh, D.N.: End-to-end information extraction by character-level
  embedding and multi-stage attentional u-net. In: BMVC. p.~96 (2019)

\bibitem{devlin2018bert}
Devlin, J., Chang, M.W., Lee, K., Toutanova, K.: Bert: Pre-training of deep
  bidirectional transformers for language understanding. arXiv preprint
  arXiv:1810.04805  (2018)

\bibitem{hong2021bros}
Hong, T., Kim, D., Ji, M., Hwang, W., Nam, D., Park, S.: {\{}BROS{\}}: A
  pre-trained language model for understanding texts in document (2021),
  \url{https://openreview.net/forum?id=punMXQEsPr0}

\bibitem{hwang2020spatial}
Hwang, W., Yim, J., Park, S., Yang, S., Seo, M.: Spatial dependency parsing for
  semi-structured document information extraction (2020)

\bibitem{li2019dice}
Li, X., Sun, X., Meng, Y., Liang, J., Wu, F., Li, J.: Dice loss for
  data-imbalanced nlp tasks. arXiv preprint arXiv:1911.02855  (2019)

\bibitem{Liu2019}
Liu, X., Gao, F., Zhang, Q., Zhao, H.: {Graph Convolution for Multimodal
  Information Extraction from Visually Rich Documents}  (mar 2019),
  \url{https://arxiv.org/abs/1903.11279 http://arxiv.org/abs/1903.11279}

\bibitem{Luong2015EffectiveAT}
Luong, T., Pham, H., Manning, C.D.: Effective approaches to attention-based
  neural machine translation. ArXiv  \textbf{abs/1508.04025} (2015)

\bibitem{park2019cord}
Park, S., Shin, S., Lee, B., Lee, J., Surh, J., Seo, M., Lee, H.: Cord: A
  consolidated receipt dataset for post-ocr parsing  (2019)

\bibitem{Qian2018}
Qian, Y., Santus, E., Jin, Z., Guo, J., Barzilay, R.: {GraphIE: A Graph-Based
  Framework for Information Extraction}  (2018),
  \url{http://arxiv.org/abs/1810.13083}

\bibitem{10.1145/1597735.1597754}
Tomanek, K., Hahn, U.: Reducing class imbalance during active learning for
  named entity annotation. In: Proceedings of the Fifth International
  Conference on Knowledge Capture. p. 105–112. K-CAP '09, Association for
  Computing Machinery, New York, NY, USA (2009). \doi{10.1145/1597735.1597754},
  \url{https://doi.org/10.1145/1597735.1597754}

\bibitem{Vedova2019}
Vedova, L.D., Yang, H., Orchard, G.: {An Invoice Reading System Using a Graph
  Convolutional Network}  \textbf{2},  434--449 (2019).
  \doi{10.1007/978-3-030-21074-8}

\bibitem{Xu2020LayoutLMv2MP}
Xu, Y., Xu, Y., Lv, T., Cui, L., Wei, F., Wang, G., Lu, Y., Florencio, D.,
  Zhang, C., Che, W., Zhang, M., Zhou, L.: Layoutlmv2: Multi-modal pre-training
  for visually-rich document understanding. In: Proceedings of the 59th Annual
  Meeting of the Association for Computational Linguistics (ACL) 2021 (August
  2021)

\bibitem{xu2020layoutlm}
Xu, Y., Li, M., Cui, L., Huang, S., Wei, F., Zhou, M.: Layoutlm: Pre-training
  of text and layout for document image understanding. In: Proceedings of the
  26th ACM SIGKDD International Conference on Knowledge Discovery \& Data
  Mining. pp. 1192--1200 (2020)

\bibitem{xu2021layoutxlm}
Xu, Y., Lv, T., Cui, L., Wang, G., Lu, Y., Florencio, D., Zhang, C., Wei, F.:
  Layoutxlm: Multimodal pre-training for multilingual visually-rich document
  understanding (2021)

\bibitem{Yu2020PICKPK}
Yu, W., Lu, N., Qi, X., Gong, P., Xiao, R.: {PICK}: Processing key information
  extraction from documents using improved graph learning-convolutional
  networks. In: 2020 25th International Conference on Pattern Recognition
  (ICPR) (2020)

\end{thebibliography}
\end{document}